\documentclass[10pt, twocolumn]{article}
\usepackage{latex8}
\usepackage{times}
\usepackage{epsfig}
\usepackage{graphicx}
\usepackage{url}
\usepackage{amsmath}
\usepackage{hyphenat}

\begin{document}

\title{A Parallel Framework for Parametric Maximum Flow \\Problems in Image Segmentation}
\author{Vlad~Olaru$^1$ \hspace{2cm} Mihai~Florea$^1$ \hspace{2cm} Cristian~Sminchisescu$^{2,1}$\\
{\em $^1$Institute of Mathematics of the Romanian Academy}, {\em $^2$Lund University}\\
{\em vlad.olaru@imar.ro, mihai.florea@my.fmi.unibuc.ro, cristian.sminchisescu@math.lth.se}\\
}
\maketitle
\thispagestyle{empty}
\setcounter{secnumdepth}{4}


\begin{abstract}

{\it This paper presents a framework that supports the implementation of parallel solutions
for the parametric maximum flow computational models widely used in image segmentation algorithms. 
The framework is based on supergraphs, a special
construction combining several image graphs into a larger one, and works on
various architectures (multi-core or GPU), either locally or remotely in a
cluster of computing nodes. The framework can also be used for performance
evaluation of parallel implementations of maximum flow
algorithms. We present the case study of a state-of-the-art image
segmentation algorithm based on graph cuts, Constrained Parametric Min-Cut
(CPMC), that uses the parallel framework to solve parametric maximum flow
problems, based on a GPU implementation of the well-known push-relabel algorithm. Our results indicate that real-time implementations based on the proposed techniques are possible.

}
\end{abstract}

\vspace*{-0.3in}
\section{Introduction}\label{sec:intro}
\vspace*{-0.1in}

\begin{figure}[htb]
        \centering \leavevmode
$
\begin{array}{ccc}
        \includegraphics[angle=0,height=0.25\columnwidth, width=0.3\columnwidth]{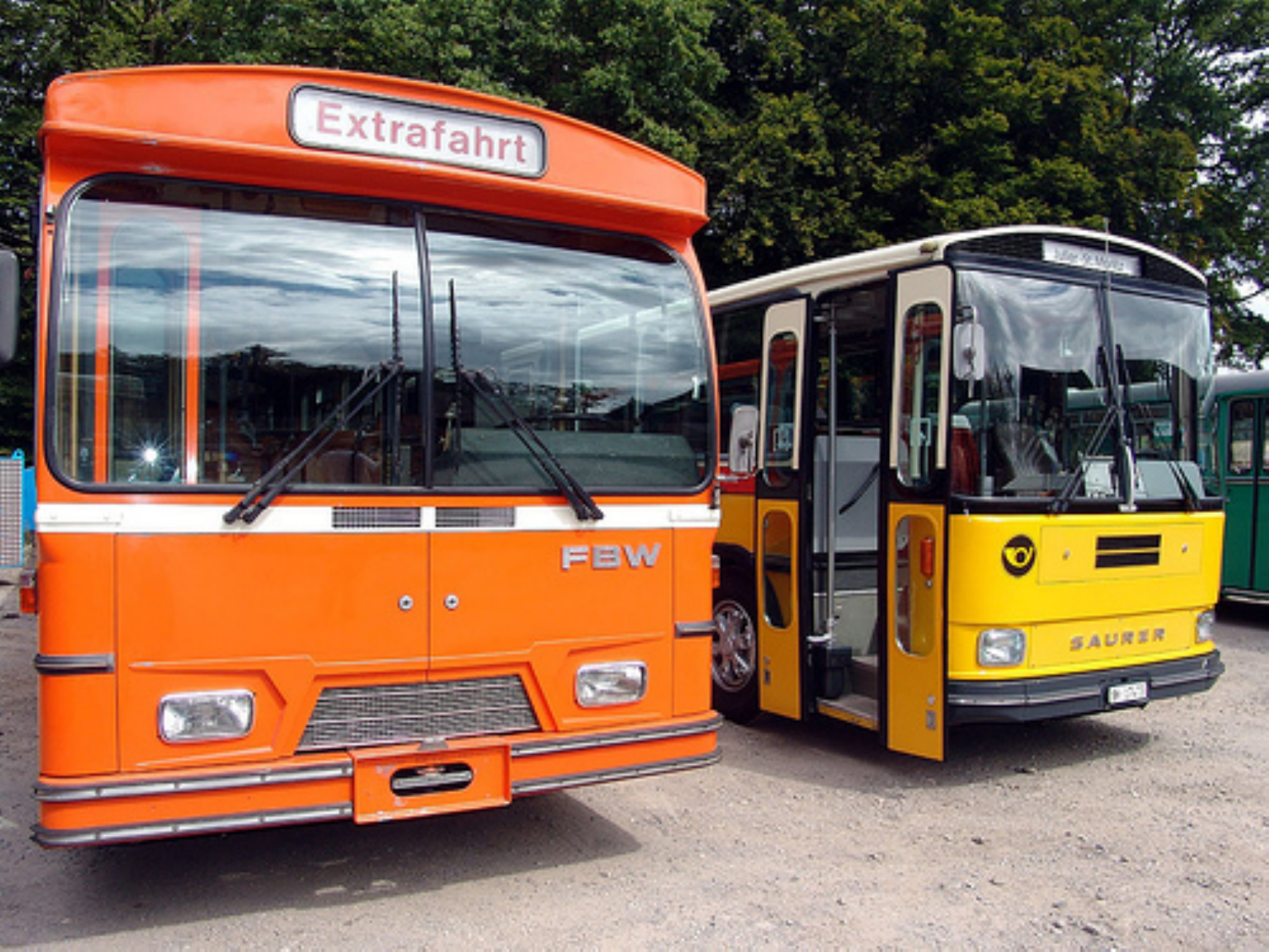} &
        \includegraphics[angle=0,height=0.25\columnwidth, width=0.3\columnwidth]{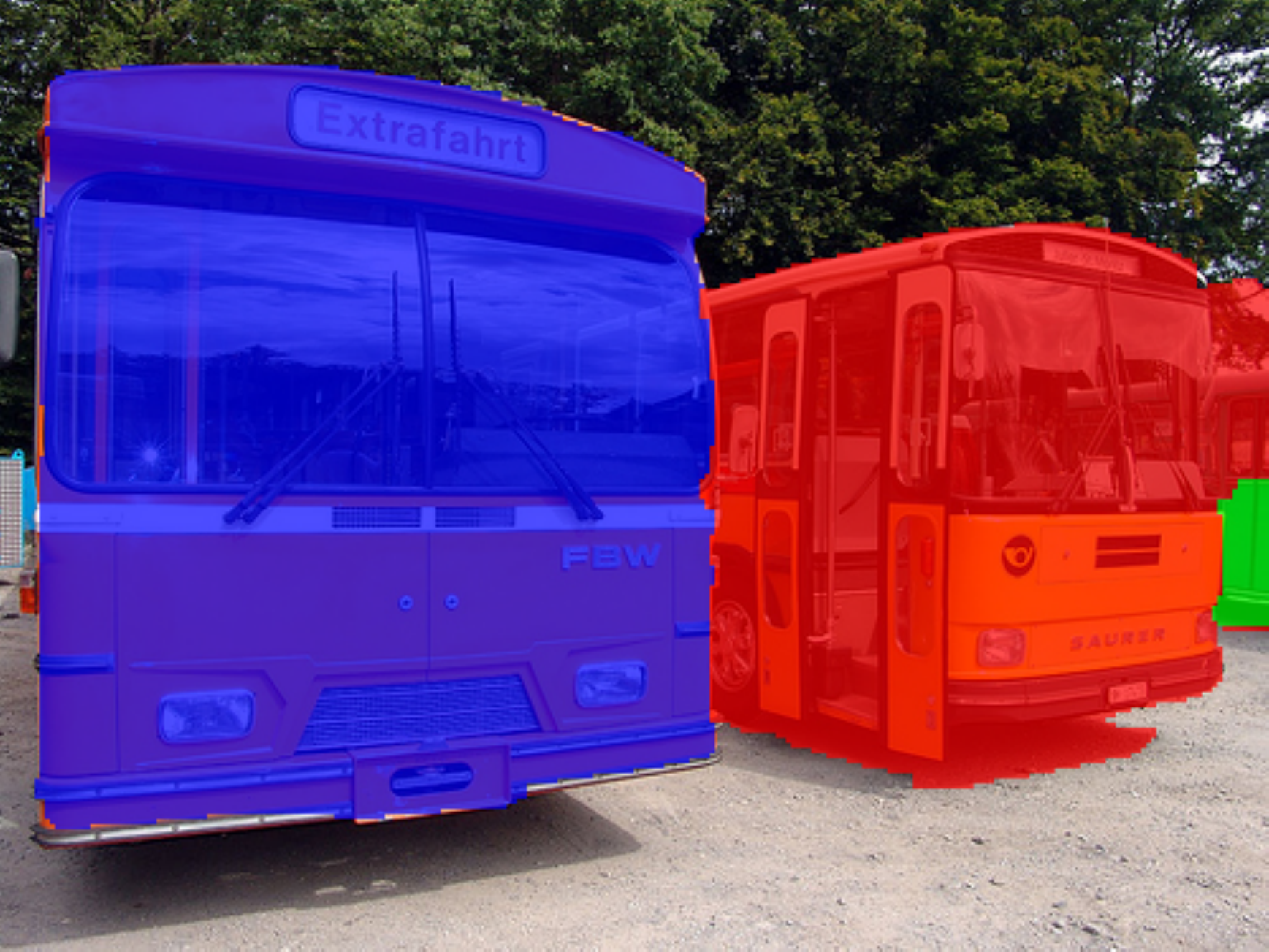} &
        \includegraphics[angle=0,height=0.25\columnwidth, width=0.3\columnwidth]{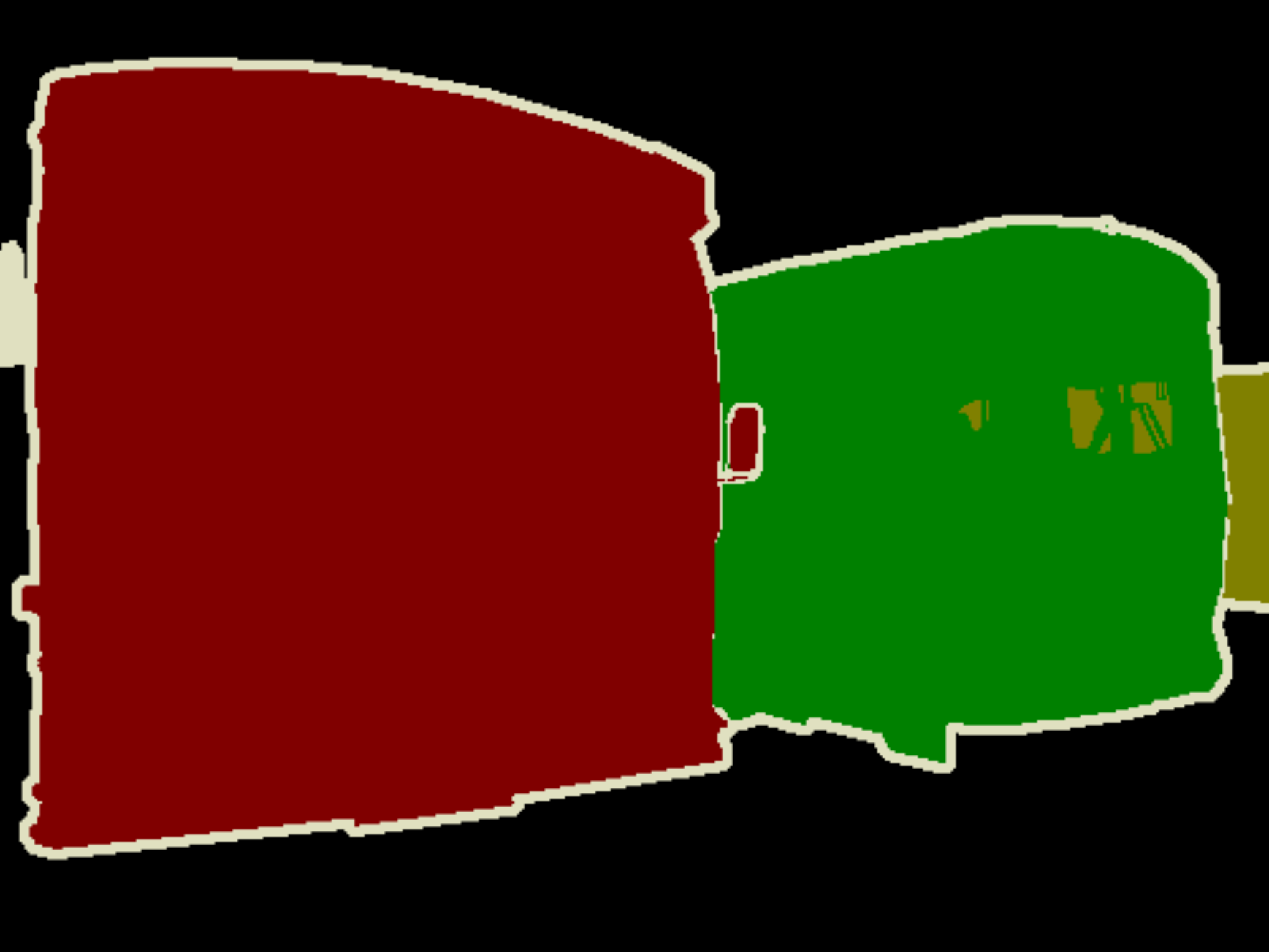}
\end{array}$
\caption{\textrm{(best seen in color) Original image, pool of segments generated by a max-flow segmentation algorithm, and ground truth respectively.}}
\vspace*{-0.2in}
\label{fig:segm}
\end{figure}

Recent advances in image segmentation \cite{CPMC} have led to improved accuracy over large and diverse image datasets \cite{VOC, H36M},
by almost doubling the performance figures. This
development has spurred the interest for the widespread use of image
segmentation models (figure \ref{fig:segm}) as a component for key tasks in computer
vision, such as video segmentation, large-scale applications for recognition
and classification or mobile computing. In this context, of particular
importance becomes the real-time performance of image segmentation
algorithms. Although reliant on advanced methodology and data structures, the
running times of the best performing algorithms still lag
behind real-time, taking a few minutes for usual images, on average.

The most advanced image segmentation algorithms involve repeatedly solving
multiple maximum-flow problems over monotonic schedules of parameter scales
(parametric max-flow) constrained at image ``seeds'' corresponding to different
locations in an image. Each image can be represented as a graph, where each
pixel is a node connecting locally with spatially adjacent ones (e.g. up, down,
left and right), and connection strengths are modulated by pixel intensity
similarity, or the presence of image contours. Solving each max-flow problem
for one setting of the parameters is equivalent to computing a binary partition
on the image graph. Performed systematically, at different locations and for
monotonic schedules of parameters, it has been empirically observed that the
process generates multiple binary segmentation hypotheses with good spatial
overlap with the different objects and scene structures present in images (see
figure \ref{fig:segm}). Often, the hypothesis generation is initiated from
different seeds independently, suggesting an
inherently high degree of parallelism. Therefore, a trivially parallel
implementation that generates solutions by running parametric maximum flow
\cite{GALLO, HOCH} independently for each seed seems appropriate.

\vspace*{-0.1in}
\vspace*{-0.07in}

However, the high computational cost of generating segment hypotheses once
a location (seed) has been selected suggests parallelizing the parametric
maximum flow procedure as an alternative way to speed up image
segmentation. Currently, as far as we know, there are no available parallel
implementations of a parametric maximum flow algorithm. In this paper, we
present the design of a general framework that can
use existing parallel graph cut solutions such as GridCut \cite{Gridcut} or
CUDA NPPI \cite{NPPI_GC, NPPI_GC_DOC} to implement a parallel algorithm that
approximates parametric maximum flow behavior. To this end, we use supergraphs,
a special construction that combines several image graphs, each having edge
weights (or capacities) that depend on a different parameter, into a larger one.

The framework is general in terms of the architectures it can
use. Supergraphs can run on multi-core processors or GPU boards, either
locally or distributed in a cluster. A parametric maximum flow problem encoded
as a collection of supergraph cut problems can be scheduled dynamically to run
on a heterogeneous collection of computing nodes. The dynamic scheduler 
efficiently adapts not only to the imbalances induced by the heterogeneous
architectures used, but also to those intrinsic to the problem, as each problem
takes a different amount of time, depending on the image complexity.

The paper also presents a case study of a state-of-the-art image segmentation
algorithm, Constrained Parametric Min-Cut (CPMC) \cite{CPMC}, that uses the
parallel framework with NVIDIA's GPU implementation of the well-known
push-relabel maximum flow algorithm \cite{Goldberg}. The comparison to a
CPMC solution based on a sequential pseudoflow algorithm \cite{HOCH} in a
trivially parallel setup (where instances of a segmentation problem
are executed independently on several cores of a processor) helps us understand how
close can we get to a real-time solution for image segmentation.

To summarize, the paper has three main contributions: {\bf (1) a parallel
solution for parametric maximum flow problems based on supergraphs} that can be
used in image segmentation algorithms; {\bf (2) a general, parallel framework for parametric maximum-flow 
problems} that can {\it (a) handle various hardware architectures}, multi-core
or GPU, both locally and remote in a cluster, {\it (b) efficiently schedule
problems} to achieve improved segmentation times, and
{\it (c) act as a performance evaluation tool} by allowing the use of various
implementations of parallel maximum flow algorithms; and
{\bf (3) a case study for a state-of-the-art image segmentation algorithm
(CPMC)}.

The paper is organized as
follows: \S\ref{sec:cpmc} describes shortly previous work on graph cuts and
image segmentation for a clear understanding of the
concepts.
In \S\ref{sec:pmf}, we present the framework solutions to parallelize the parametric
maximum flow algorithms, using supergraphs. \S\ref{sec:evaluation} presents the results of our case study evaluation of CPMC. We conclude in \S\ref{sec:conclusions}.

\vspace*{-0.1in}
\section{Graph-Cut based Image Segmentation}\label{sec:cpmc}
\vspace*{-0.1in}

Graph cuts can be used to segment an image into a foreground object and the
rest of the image, usually referred to as background, in order to obtain a figure-ground segmentation. This is a form of binary
classification, with 1 assigned to foreground pixels and 0 to background.

\begin{figure}[htb]
\scalebox{0.32}{
        \includegraphics[angle=0,viewport=0.01cm 11cm 18cm 25.6cm, clip=true]{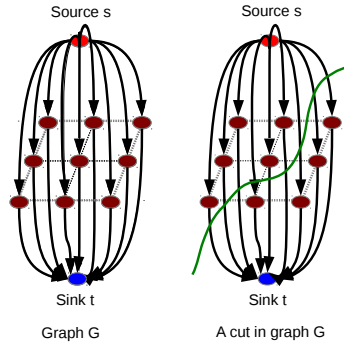}}
\vspace*{-0.2in}
\caption{\textrm{Associated image graph and a cut example (best seen in colors).}}
\vspace*{-0.2in}
\label{fig:gc}
\end{figure}

The binary inference (labeling) process is performed by running a maximum flow/minimum cut
algorithm on a graph whose vertices represent the pixels in the image. Two
special vertices, the {\it source s} and the {\it sink t} are connected to every
vertex of the graph by means of weighted edges (see figure \ref{fig:gc});
the weights are called edge capacities. For image segmentation,
the source and sink are associated with the two labels that will be used to
distinguish the foreground object from the background. The weights of the edges
that link {\it s} and {\it t} to the graph vertices (the pixels) quantify a
penalty expressing how correct is to assign that pixel to either of the two
classes of labels represented by the source and the sink.

\vspace*{-0.1in}

Regular graph vertices (corresponding to image pixels) are linked to each
other by weighted edges as well. Typically, image segmentation models use the
weights of the edges that connect each vertex to its nearby neighbors (up, down
and laterally) to model smoothness, i.e. the assumption that nearby
pixels are likely to have similar labels. 

An {\it s-t cut} of the graph is a partitioning of the vertices into two
disjoint subsets: one containing vertex {\it s} and the other one containing
vertex {\it t}. The {\it cost} of the cut is defined to be the sum of the
weights of those edges in the graph that have one vertex in the s-partition
and the other in the t-partition. A {\it minimum cut} corresponds to those
graph cuts that have minimum cost.

A graph cut induces a labeling of the image pixels, depending on which
partition they were inferred to. The problem of finding a cut is equivalent to the one of minimizing an energy defined on the graph. The energy has two terms, depending on which type
of edges the cut crosses: edges linking either {\it s} or {\it t} to a regular
vertex (pixel), or regular edges that link neighboring pixels. The first
category of terms is called ``data'' or ``unary'' terms, while the second
accounts for the ``pairwise'' terms (regularization terms). A minimum cut in such an image graph corresponds to a minimum energy among all
of the possible label configurations of the image graph. 

Greig et al. \cite{Greig} have used this method for the first time to smooth
noisy images and showed that the maximum a posteriori estimate of
a binary image corresponds exactly to the maximum flow in the associated image
graph constructed as previously described. According to the Ford and
Fulkerson theorem \cite{FF}, a maximum flow from {\it s} to {\it t} {\it saturates} the sum
of the capacities of a set of edges in the graph that partitions the vertices
into two disjoint sets that actually correspond to a minimum cut in the graph.

There are many polynomial time algorithms that solve the maximum flow problem
(see \cite{Cook}), including augmenting path (Ford-Fulkerson \cite{FF}) and
push-relabel \cite{Goldberg} algorithms, but their presentation is beyond the
scope of this paper. An augmenting path algorithm
widely used in computer vision is due to Boykov and Kolmogorov \cite{Boykov}. An
extended view on the use of graph cuts in computer vision can be found in
\cite{BVeks}.

\vspace*{-0.1in}
\begin{figure}[htb]
\begin{center}
\scalebox{0.3}{
        \includegraphics[angle=0,viewport=0.1cm 14cm 20cm 26.5cm, clip=true]{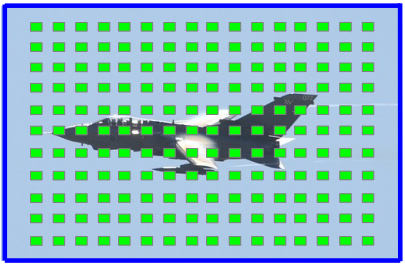}}
\vspace*{-0.2in}
\caption{\textrm{(best seen in color) A regular grid of seeds in an image. Binary partitions (segmentations) are extracted around regularly placed foreground seeds (green dots) that express the foreground bias, while background seeds (in blue) are placed usually on the border of the image. Each generated segment corresponds to a graph cut segmentation problem.  For an entire image, multiple independent solutions are generated at the cost of heavy graph cut computations.}}
\vspace*{-0.25in}
\label{fig:seeds}
\end{center}
\end{figure}

\vspace*{-0.25in}
\subsection{Parametric Max Flow \& Image Segmentation}\label{sec:pmf-desc}
\vspace*{-0.1in}

Parametric max flow algorithms \cite{GALLO,Kolmogorov} are used in image
segmentation to generate a set of hypotheses for plausible object segments
in a given image. They are able to optimize
energies where the unknowns are both the binary labels of pixels and the
weighting (scale) $\lambda$ between the unary and pairwise terms of the energy
model. The $\lambda$ values for which the corresponding energy value
changes are called ``breakpoints'' and mark the optimal
solutions of a parametric max flow problem. In the ``monotonic'' case where the factors multiplying the parameter $\lambda$ in
the unary (data) energy terms are all non-negative or non-positive, the optimal
solutions are nested \cite{GALLO} and an efficient implementation of the
parametric maximum flow algorithm is possible. The algorithms can either
compute all breakpoints (an upper bound is the number of the graph nodes) or
a subset of them. Either way, monotonicity makes the calculation
significantly more efficient as earlier computations are reused. In practice,
a preset list of parameter values (usually defined on a logarithmic scale),
the so called {\it $\lambda$-schedule}, can be used instead of computing all the
breakpoints, as empirical evaluations \cite{CPMC} have shown that the ground
truth covering stays almost the same, at significantly lower computational cost
due to the reduced number of breakpoints generated (and thus, a reduced number
of segment hypotheses). We say that this type of run ``approximates''
parametric max flow behavior.

Graph cut problems (preferably monotonic) are associated with different seeds in order to generate a pool of segments with high probability of (foreground) object
overlap. A seed is a set of pixels ``frozen'', by construction, to belong to either
foreground or background. The foreground seeds are usually placed regularly on a
grid in the image, whereas the background seeds are assigned on the borders of
the image (see figure \ref{fig:seeds}). A collection of maximum flow problems is
solved for each pair of foreground and background seeds and different $\lambda$
values (the $\lambda$-schedule), that are used to express the so called
foreground bias associated with the non-seed pixels.
The result is a large and diverse set of segments of
different sizes and structural (shape) relevance.

\vspace*{-0.1in}
\subsection{Trivial Parallelism on Multi-Core Processors}\label{sec:multicore}
\vspace*{-0.1in}

A list of problems defined by a pair of foreground and background seeds
and a $\lambda$-schedule can be solved independently on different
processing units, given that no two pairs of foreground and background
seeds are the same. For instance, a trivially parallel solution can be
implemented by using MATLAB's {\it parfor} instruction (or similar instructions
in other languages) that executes each iteration independently as a thread on
one of the available processor cores.
\vspace*{-0.1in}
\begin{figure}[htb]
\scalebox{0.55}{
        \includegraphics[angle=0,viewport=2.5cm 19cm 18cm 27cm, clip=true]{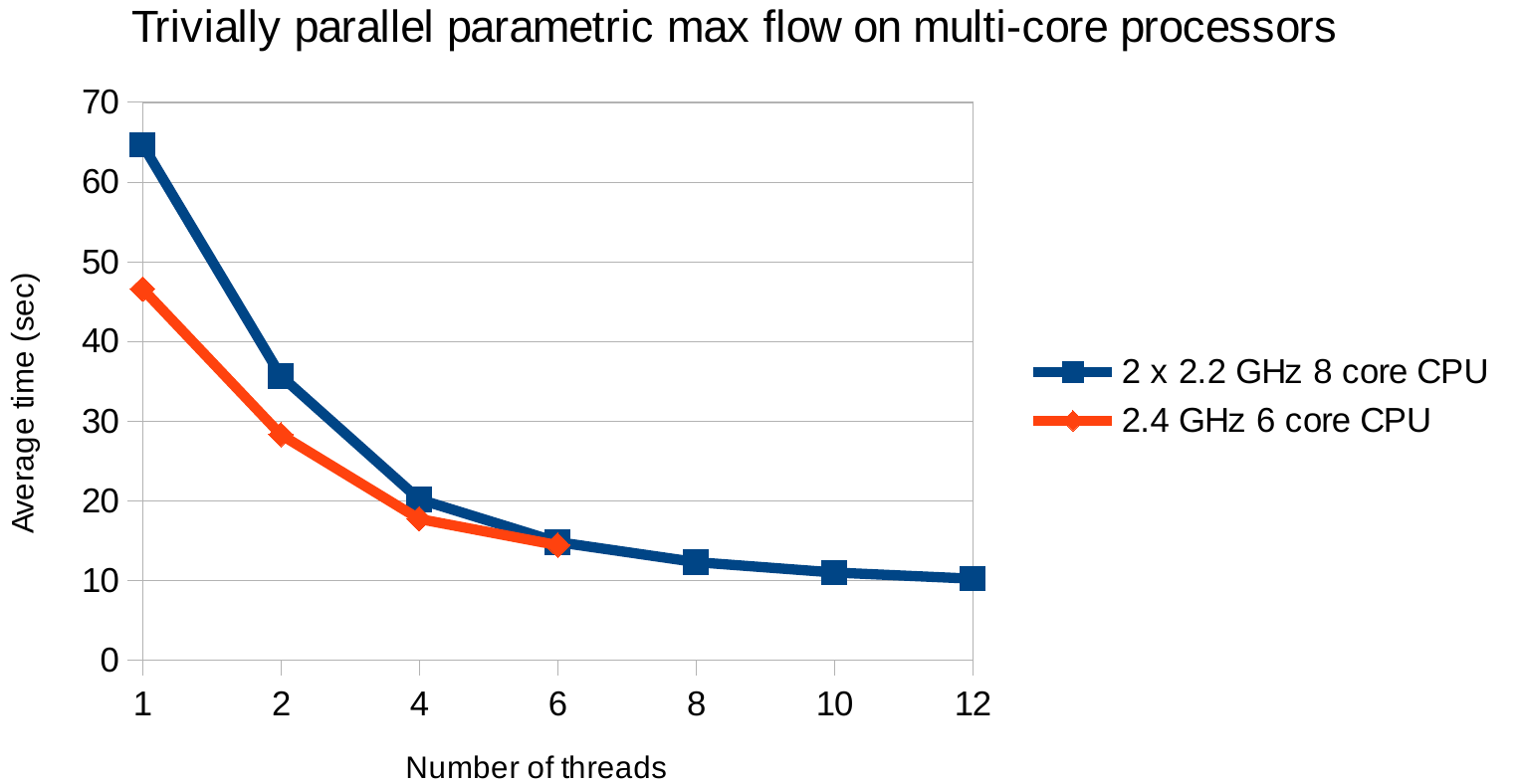}}
\caption{\textrm{Trivial parallelism performance for figure-ground segmentations.}}
\label{fig:tp-bound}
\end{figure}
\vspace*{-0.28in}
The main advantage of this type of parallelization is simplicity, both in
terms of programming effort and negligible need of synchronization of the
worker threads (thus enabling maximum parallelism). From a programming
standpoint, one only needs to mark the
appropriate {\it parfor} code blocks. The programming model
is sequential for any particular {\it parfor} code block solving a problem, and
the speed-up comes from the high usage of the available cores of the
processor.

However, this parallel solution does not attempt to speed up any of the
individual problems, which run sequentially.
Figure \ref{fig:tp-bound} shows the mean time in seconds taken by the
pseudoflow algorithm \cite{HOCH} to yield figure-ground hypotheses for a set
of 500 images from the VOC  \cite{VOC} dataset, with a schedule of 20 $\lambda$
values and 178 seeds, each.  Please note that, in practice, even with a relatively
small number of processing units (cores), the speed-up of the trivially
parallel solution flattens out quite quickly. As soon as 10 cores are used, the
performance of the slower processor (Intel Xeon E5-2660) starts to saturate
above 10 seconds, still far from real-time expectations. Even on the
faster processor (Intel Xeon E5-2620 v3) the execution times get close to the
other processor times as soon as 6 cores are used. This lack of scalability
motivates the need to investigate parallel solutions for the parametric maximum
flow solver as well.

\vspace*{-0.1in}
\section{Parallel Parametric Max Flow Solution}\label{sec:pmf}
\vspace*{-0.1in}

To the best of our knowledge, there is no available parallel implementation of
a parametric maximum flow algorithm. A few prior articles focus on the topic of
parallel implementations of maximum
flow \cite{Hussein, Vineeth}, but don't offer code. One available
implementation is GridCut \cite{Gridcut}, and works for multi-core
processors only. It defines a grid of computing units that can process graph
cuts in parallel based on a popular augmenting path algorithm featuring
tree-reuse \cite{Boykov}. GridCut implements adaptive bottom-up merging
\cite{ABM} and cache efficient memory layout \cite{CacheEffGC}. Other
available implementations of max flow algorithms are GPU implementations
\cite{NPPI_GC, NPPI_GC_DOC, CMF-TR, CMF}. The NVIDIA NPPI library
\cite{NPPI_GC,  NPPI_GC_DOC} implements a push-relabel algorithm
\cite{Goldberg}. 

Given the circumstances, running a parallel parametric maximum flow algorithm
proves challenging. One solution would be to seek an ``approximation'' of
the parametric behavior (in the sense defined in \S\ref{sec:pmf-desc}) by
using a preset $\lambda$-schedule, and run a parallel maximum flow routine
once for each $\lambda$ value in the schedule. However, this ``batch'' call is far
from optimal, since it is proven that, in a monotonic case, a parametric
maximum flow algorithm can run asymptotically close to a regular maximum flow
algorithm \cite{GALLO}, i.e.  with the same theoretical
complexity. Hence, this batch procedure is a poor match to what an
optimal parallel parametric maximum flow algorithm could achieve in theory.

\vspace*{-0.1in}
\subsection{Parametric Max Flow with Supergraphs}\label{sec:supergraphs}
\vspace*{-0.1in}

Besides the inability to optimize computations in the monotonic case,
another shortcoming of the batch method is that running a single graph cut
problem at the time, either on a multi-core processor or a GPU architecture,
might not use the available hardware resources to the fullest. This becomes
striking especially for the latest generation of GPU boards that feature
thousands of computing cores. 

To address the issue, consider running several graph cut problems
simultaneously on the available parallel computing infrastructure. This is not
straightforward, since the programming interfaces of parallel graph
cut routines supplied by software like GridCut or CUDA NPPI take a single graph as parameter. The solution is to ``knit'' together several graphs representing
different problems into a larger graph, that we call a ``supergraph'', and to
pass it on, as a parameter to the graph cut calls.

These supergraphs represent the building block of our parallel framework for
parametric max flow problems in image segmentation and can be constructed at
two levels: $\lambda$ and seed level. A $\lambda$-supergraph combines together
graphs for several $\lambda$ values, whereas a seed supergraph combines several
$\lambda$-supergraphs together. Usually, our structures combine an entire $\lambda$-schedule, the list of the
$\lambda$s that we run the parametric max flow with, but it is possible to have
smaller supergraphs as well. In the case of seed supergraphs though, we use only
$\lambda$-supergraphs constructed for an entire $\lambda$-schedule.

\begin{figure}[htb]
        \centering \leavevmode
\scalebox{0.31}{
        \includegraphics[angle=0,viewport=0.5cm 10.7cm 22cm 23.9cm,clip=true]{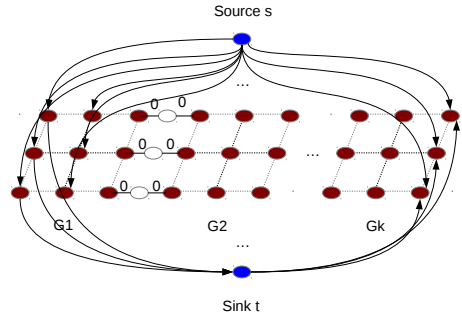}}
\vspace*{-0.1in}
\caption{\textrm{Supergraph composed out of {\it k} individual graphs {\it $G_1$, $G_2$, \ldots  $G_k$} (best seen in colors).}}
\label{fig:supergraph}
\end{figure}

Combining two graphs into a supergraph can be simply done by inserting
additional vertices ``between'' the two graphs and by linking them to the
regular vertices from the left and right graph by means of zero-weight edges
(see figure \ref{fig:supergraph}).
Inductively, one can build arbitrarily large supergraphs out of
individual graphs. Any minimum cut in a supergraph built like that is a union
of the disjoint minimum cuts of the original graphs knitted together, plus
some zero-weight edges that do not count towards the overall cost of the
supergraph cut. Therefore, computing minimum energies associated with a pair of
foreground-background seeds and a given $\lambda$ value can be derived from
such a supergraph by decomposing the minimum supergraph cut into its individual
minimum cut components. In other words, computing a supergraph maximum
flow/minimum cut approximates the behavior of a parallel parametric maximum
flow algorithm running on the individual graphs.

\vspace*{-0.1in}

\vspace*{-0.25in}
To see why this works, consider the following situation. Let {\it S} be
a supergraph composed of {\it k} individual graphs {\it $G_1$, $G_2$, \ldots
$G_k$} (see figure \ref{fig:supergraph}). Let's assume {\it C} is a minimum
cut in {\it S}, and {\it $C_1$, $C_2$, \ldots  $C_k$} are the minimum cuts of
the individual graph components. Suppose that one of the individual cuts, say
{\it $C_i$}, is not a minimum cut in {\it $G_i$}. Then, there is a minimum
cut {\it $C'_i$} in {\it $G_i$}, different than {\it $C_i$}. Since {\it $G_i$}
is linked by means of zero weight edges to its neighboring graphs in {\it S},
any minimum supergraph cut that crosses {\it $G_i$} needs to include
{\it $C'_i$}, because the zero-weight crossing edges do not contribute to the
overall cost, {\it $C'_i$} is the minimum cost path severing {\it $G_i$} into
two partitions and the supergraph cut must somehow cross {\it $G_i$}. Therefore,
there is another supergraph cut including {\it $C'_i$} of smaller cost than
{\it C}, which contradicts the assumption that C is a minimum supergraph
cut. Thus, all of the {\it $C_is$} must be minimum cuts in their
corresponding {\it $G_i$s}.

Conversely, suppose that there is a supergraph cut {\it C'} in {\it S} whose
cost is smaller than that of {\it C}. Then, there is another union of
individual minimum cuts {\it $C'_1$ $\cup$ $C'_2$ $\cup$ \ldots $\cup$ $C'_k$}
that compose {\it C'}. Since {\it $C'_i$}s are disjoint sets, it means that at least one of them from {\it C'}, say
{\it $C'_j$}, is smaller than its corresponding {\it $C_j$}, which contradicts
the assumption that {\it $C_j$} is a minimum cut in {\it $G_i$}.

Therefore, it is sound to use this procedure to amass several graphs into
a supergraph and use the individual components of the minimum supergraph
cut as individual minimum cuts. Thus, one can solve simultaneously either a
single seed problem (by building a $\lambda$-supergraph) or several seed
problems (by binding together several $\lambda$-supergraphs for several
seeds). The ability to run custom sized graphs ends up in a better usage of the
available computing power of the underlying hardware architecture.


We have used supergraphs both with GridCut and CUDA NPPI, but our CPMC case
study focuses on the GPU solution, as our evaluations showed that GridCut
performs significantly worse than CUDA. Nevertheless, we emphasize that the supergraph method is general and can be used with any available
parallel graph cut implementation as a means to compute a parametric maximum
flow in parallel. The method is especially effective when the parallel graph cut
source code is not available (e.g. CUDA NPPI).

\vspace*{-0.1in}
\subsection{Exposing Additional Supergraph Parallelism}\label{sec:swap}
\vspace*{-0.1in}

It is known that exchanging the roles of source and sink, operation that we
call an {\it s-t swap}, does not affect the results of a graph cut algorithm
(i.e., the maximum flow/minimum cut remain the same). However, it might help
a parallel implementation of a push-relabel algorithm, like NVIDIA's
{\it nppiGraphcut} \cite{NPPI_GC, NPPI_GC_DOC}, run faster \cite{Stich}. The
reason for this behavior is that the parallel workload at every iteration of
the algorithm is given by the number of regular vertices (pixel nodes) that
have residual capacity on their edges to/from source independent of the edges
to/from sink \cite{Stich}. So, if there are more such vertices for the sink
than for the source, swapping them exposes more parallelism.

Choosing the source and the sink by running the algorithm twice, once with the
original graph and then after an {\it s-t swap}, to see which run yields faster
results, is obviously not a solution. Instead, we use an heuristic
to choose the source and sink. For each regular vertex in the graph,
the difference between the capacities of its source and sink edges is
computed. Then, for each vertex, we separately add the positive and negative
differences. If the number of negative differences turns out to be larger than
that of positive ones, we apply the {\it s-t swap}.  This procedure leads to more hardware resources active per algorithm iteration and
improves performance significantly (see \S\ref{sec:basicGPU}).

When using {\it s-t swaps} with supergraphs, one has to properly choose the source and sink so that every individual graph is
aligned for maximum available parallelism. Thus, all the individual
graphs composing a supergraph must be checked if they need to be
{\it s-t swapped} so that the resulting supergraph has a source and
a sink that allow the highest possible degree of parallelism. That is easier
done for $\lambda$-supergraphs, because such a supergraph represents the
same problem (i.e., the same foreground-background pair of seeds), but care
must be taken when building seed supergraphs, that might need to reverse some
of the individual $\lambda$-supergraphs.  

\vspace*{-0.1in}
\subsection{Using the Parallel Framework}\label{sec:cluster}
\vspace*{-0.1in}

A collection of seed problems, each represented by a pair of foreground and
background seeds and a $\lambda$-schedule, is going to be encoded by means of
supergraphs, as previously described. The resulting set of supergraphs
can have a smaller size than the collection of seed problems if several
such problems are expressed by means of seed supergraphs. Each resulted
supergraph gets scheduled for parametric maximum flow processing on a given 
computing node (also called server), either locally or remote. Remote
processing is achieved by means of Remote Procedure Calls (RPC) for the
supergraph cut routines. The scheduling is controlled by a master node,
which runs the image segmentation algorithm. The master
node can act as computing server as well, but in this case the local
computing architecture, either CPU- or GPU-based, is going to be accessed
directly instead of performing an RPC. The resulting cluster of servers collaborating to solve the collection of seed problems may be
heterogeneous, regardless whether operation is on CPUs or GPUs.

\vspace*{-0.15in}
\subsubsection{Supergraph Scheduling}\label{sec:sched}
\vspace*{-0.05in}

The master node can perform two types of parallel, non-preemptive scheduling:
static and dynamic. Static scheduling assigns supergraphs to computing servers
by using a MATLAB {\it parfor} instruction with {\it n} threads in which each
parallel loop iteration {\it i} gets allocated task {\it i mod n}. All the
tasks allocated to a class {\it i mod n}, e.g., those that execute an
RPC to a given remote server, will be executed sequentially and non-preemptively,
one after the other. Thus, the makespan, the maximum value among the completion
times of the tasks, will be determined by the time needed to run the longest
class {\it i} of tasks {\it i mod n}.

However, in a heterogeneous computing environment with different hardware
architectures (for instance, different types of GPU boards, as in our case
study), significant computational load imbalances may arise. Even the
same hardware, say two GPU boards of the same kind, will not yield the same
performance when accessed locally vs. over the network via RPC. Moreover, there
is an intrinsic source of imbalance in the segmentation problem, because
different seeds of an image induce different image graphs and therefore
different graph cut computational costs.

The dynamic scheduler, which is also multi-threaded, attempts to offset these
load
imbalances by picking up a server from a list of available ones in FIFO order
and executing the RPC (or a local call, in the case of the master node) to that
node with a supergraph as a parameter. The server is removed from the list and,
later on, when the supergraph cut call finishes, is inserted back into the
list. Hence, the list of available servers grows and shrinks dynamically and,
at times, may become empty, in which case no server is available for computation
and the master program dispatching the tasks gets blocked. 

In contrast to static scheduling, this dynamic policy that handles
a supergraph as soon as a server is available offers a more balanced mix
of task overlaps, which in turn should contribute to a smaller makespan. Optimal
scheduling of independent, non-preemptable tasks to minimize the makespan is
known to be NP-hard \cite{HBookS}. However, a priori knowledge about the
supergraph cut processing times may improve the worst-case performance of
the scheduling algorithm. For instance, sorting the task list in non-increasing
order of processing times before scheduling and assigning the first available
task to the first available server during scheduling, policy known as
{\it Largest Processing Time First (LPT)}, is an effective way to minimize the
worst-case makespan \cite{HBookS}. In our case, this a priori information is hard to
get though, because it is highly data dependent.


The scalability of the framework depends on two main components: the scalability
of the master scheduler and that of the parallel graph cut routine processing
the supergraphs. For the latter, our framework is limited by the scalability of
the available software used (i.e., GridCut, CUDA NPPI, etc). The scalability of
the scheduler is influenced by the available resources on the master
node and the network latency for remote communication. However, as our
experiments show (see \S\ref{sec:cluster5} and \S\ref{sec:cluster5-10}),
achieving near real-time
performance doesn't necessarily assume many computing nodes. Therefore, one can
conclude that the master scheduler shouldn't face scalability problems as
long as it can use small-scale multiprocessor (6-8 cores) machines.

\vspace*{-0.15in}
\subsubsection{Supergraphs and Network Communication}\label{sec:comm}
\vspace*{-0.05in}

Typical sizes of the images in the VOC dataset \cite{VOC} used in our
case study amount to roughly 80-100K pixels, and so are the sizes
of the corresponding image graphs. Usually, for computational reasons, image
segmentation algorithms like CPMC downsample images to half, so the resulting
graph size sums up to approximately 160-200KB of memory (for 4-byte floats or
integers). Packing several such graphs into a supergraph can enlarge
the size of the RPC parameters even further. Moreover, library calls
like NVIDIA's {\it nppiGraphcut} \cite{NPPI_GC,  NPPI_GC_DOC} require five
such large matrices as parameters, among others. As a result, the overall size
of the RPC parameters that have to be transferred over the network tends to be
quite large and thus may have a negative impact on the performance of the call.

One possible solution to alleviate the consequences of transferring large
amounts of data over the network is to minimize the number of transfers
by packing several graphs into a larger supergraph (say, instead of sending a
single $\lambda$-supergraph parameter, one might send a two seed supergraph,
i.e. two $\lambda$-supergraphs). Thus, the overhead of the send/receive network
operations gets amortized over larger amounts of data and the transfer
performance increases.

One could also attempt to make better use of the available bandwidth by
overlapping communication with computation. Issuing two concurrent RPCs to the
same server results in an overlap of the execution of the first call with the
transfer of the parameters of the second call. Naturally, handling two
concurrent RPCs requires multi-threaded server capabilities.
Given that the TI-RPC Linux package does not include multi-threaded support for
server side RPC (unlike the original Sun Microsystems/Oracle version), we had
to implement a multi-threaded RPC server as well, but this choice turned out to
be beneficial in our case study for CPMC using NVIDIA's NPPI library (see \S\ref{sec:rpc}).

\vspace*{-0.15in}
\section{Case Study: CPMC}\label{sec:evaluation}
\vspace*{-0.1in}

The CPMC release \cite{cpmc-release} can use two parametric maximum flow
algorithms \cite{GALLO, HOCH}. In our evaluation, we have chosen the
pseudoflow algorithm \cite{HOCH} because it can also run ``approximately''
(see \S\ref{sec:pmf-desc}), i.e. without computing all the breakpoints, by
accepting as argument a preset $\lambda$-schedule. Thus, the whole CPMC
algorithm runs faster and the comparison to our framework is fair. The other
option \cite{Babenko, GALLO} works only by computing all the breakpoints online.

In this setup, CPMC iteratively solves a list of independent problems defined
by a pair of foreground and background seeds and a $\lambda$-schedule passed to
the pseudoflow algorithm. The problem solver is implemented in MATLAB,
while the pseudoflow solver is implemented in C (hooked up
with the MATLAB code by means of MEX libraries). Thus, a trivially parallel
solution can be easily implemented by using MATLAB's {\it parfor} instruction
that executes each iteration independently as a thread on one of the available
processor cores.


Motivated by the argument in \S\ref{sec:multicore}, we compared the pseudoflow
based solution to that of the supergraph framework, which parallelizes the
figure-ground stage of CPMC \cite{CPMC}, in order to assess its utility as a
tool towards real-time performance for image segmentation. To that end, we have
employed a cluster of GPUs managed by the framework as described in
\S\ref{sec:pmf}. The GPU cards have run the push-relabel
implementation of the NVIDIA NPPI library \cite{NPPI_GC, NPPI_GC_DOC}. With
no access to the library source code, we had to use the NVIDIA code as a
black box, with no possibility to perform any kind of code
optimization.

\vspace*{-0.1in}
\subsection{Evaluation Setup}\label{sec:setup}
\vspace*{-0.1in}

The evaluation has been driven on two types of HP
workstations: three Z840 stations equipped with one Intel(R) Xeon(R) CPU E5-2620
v3 @ 2.40GHz processor (6 cores) and 32GB RAM, and one Z420 station equipped with
an Intel(R) Xeon(R) CPU E5-1650 v2 @ 3.50GHz processor (6 cores) and 32 GB
RAM. We used five NVIDIA GPUs for the experiments, two Tesla K40
(one per Z840 station) and three Titan Black boards (two hosted on a Z840
machine, and the third on the Z420 station).
A Tesla K40 board features 2880 cores clocked at 745MHz
and 12 GB of RAM, while a Titan Black board has 2880 cores running at
889 MHz and 6 GB of RAM. We used CUDA 6.5 for the experiments. All the systems
run Linux and are connected by Gigabit Ethernet.

Unless otherwise stated, all the experiments use a 500 image subset of the
VOC2012 dataset \cite{VOC} and evaluate over this subset of images the minimum,
maximum, and average time values, taken by the graph cuts of the
CPMC \cite{CPMC} figure-ground segmentation stage that yields
the segment hypotheses. This stage uses three different Segmenter methods for a
total of 178 seeds and 20 $\lambda$ values for each of these seeds (these are
the default values, for details see \cite{cpmc-release}).

\vspace*{-0.1in}
\subsection{Performance of the Pseudoflow Algorithm}\label{sec:pseudoflow}
\vspace*{-0.1in}

The first experiment attempts to assess the performance of the
pseudoflow algorithm \cite{HOCH} on multi-core architectures. As already
mentioned, this is a sequential parametric maximum
flow algorithm and can be easily used with the {\it parfor} MATLAB instruction
to implement a form of trivial parallelism on multi-core architectures. This
parallel algorithm will represent our multi-core baseline performance.

\vspace*{-0.1in}
\begin{table}[htb]
\centering
\begin{tabular}{ccccc}
\hline
\hline
Time  & Min & Avg & Max \\
\hline
\hline
1 parfor thread & 13.35 s & 46.59 s & 72.59 s &   \\
\hline
2 parfor threads & 8.14 s & 28.29 s & 42.70 s &   \\
\hline
4 parfor threads & 5.35 s & 17.71 s & 26.23 s &   \\
\hline
6 parfor threads & 4.10 s &  14.42 s & 21.48 s  \\
\hline
\end{tabular}
\caption{\textrm{Parametric pseudoflow performance on multi-core architectures.}}
\label{tab:pseudoflow}
\vspace*{-0.2in}
\end{table}

\vspace*{-0.1in}
We ran CPMC with the pseudoflow solver on a 6-core Z840 workstation
with a $\lambda$-schedule of 20 values by varying the number of threads
of the {\it parfor} instruction from 1 to 6. The results are shown in table
\ref{tab:pseudoflow} and represent the average, maximum and minimum times for
performing the graph cuts of the CPMC figure-ground segmentation stage. These
numbers appear also in figure \ref{fig:tp-bound} and show that increasing
the number of cores doesn't help in the long run.

\vspace*{-0.1in}
\subsection{Basic Performance of Push Relabel on GPUs}\label{sec:basicGPU}
\vspace*{-0.1in}

Our next experiment on the VOC image subset attempts to shed some light on
the local performance of our GPU cards (that is, without RPC) when using
the methods described in \S\ref{sec:pmf}. We ran the experiments on the Z840
workstations using $\lambda$-supergraphs (i.e., one seed supergraphs) with 20
$\lambda$ values when calling the NVIDIA NPPI {\it nppiGraphcut} routine. First, we show
in table \ref{tab:batch} the performance of the GPU boards without using
supergraphs at all. This is the method we called ``batch'' in \S\ref{sec:pmf}
that calls iteratively the {\it nppiGraphCut} routine for every $\lambda$ value
in the schedule. These results are the baseline for our next comparisons.

\vspace*{-0.1in}
\begin{table}[htb]
\centering
\begin{tabular}{ccccc}
\hline
\hline
Time & Min & Avg & Max \\
\hline
\hline
Tesla K40 & 61.42 s & 140.88 s & 256.54 s &   \\
\hline
Titan Black & 45.07 s & 102.28 s & 181.38 s &   \\
\hline
\end{tabular}
\caption{\textrm{Batch performance of NVIDIA's push relabel implementation.}}
\label{tab:batch}
\end{table}

\vspace*{-0.4in}
\begin{table}[htb]
\centering
\begin{tabular}{ccccc}
\hline
\hline
Time & Min & Avg & Max \\
\hline
\hline
Tesla K40 & 17.70 s & 63.53 s & 167.16 s &   \\
\hline
Titan Black & 13.61 s & 47.47 s & 122.69 s &   \\
\hline
\end{tabular}
\caption{\textrm{Performance of NVIDIA's push relabel implementation without {\it s-t swaps}.}}
\label{tab:noswap}
\vspace*{-0.2in}
\end{table}

\vspace*{-0.1in}
Table \ref{tab:noswap} shows the performance figures of the two boards when
using supergraphs without applying the {\it s-t swap} optimization (see
\S\ref{sec:swap}). The comparison to table \ref{tab:batch} reveals that the
use of supergraphs reduces the average figure-ground segmentation latency
more than two times for both type of boards, thus making a strong case for
the use of supergraphs. Please also note the minimal latencies, where the
supergraph method yields at least three times lower figures.

\vspace*{-0.1in}
\begin{table}[htb]
\centering
\begin{tabular}{ccccc}
\hline
\hline
Time & Min & Avg & Max \\
\hline
\hline
Tesla K40 & 10.45 s & 32.80 s & 54.76 s &   \\
\hline
Titan Black & 8.40 s & 25.74 s & 42.77 s &   \\
\hline
2 x Titan Black & 4.53 s & 13.93 s & 22.77 s &   \\
\hline
\end{tabular}
\caption{\textrm{Performance of NVIDIA's push relabel implementation using properly {\it s-t swapped} supergraphs.}}
\label{tab:basicGPU}
\vspace*{-0.2in}
\end{table}

\vspace*{-0.1in}
Table \ref{tab:basicGPU} presents the results of using supergraphs that are
properly {\it s-t swapped} for optimal performance. The third row of the table
presents the results of using simultaneously both of the Titan Black cards in
one of our Z840 stations. On average, the Titan Black board takes roughly 78\%
of the K40 time and is almost 55\% faster than the trivially parallel, single threaded algorithm (see table \ref{tab:pseudoflow}). It is also worth
noting that two Titan Blacks together outperform, on average, the trivially
parallel algorithm on six threads.

The comparison to table \ref{tab:noswap} shows that the {\it s-t swap}
operation on supergraphs cuts almost to half the figure-ground segmentation
latency for Tesla K40 and by roughly 46\% for Titan Black, on average. Also
noteworthy is that the maximum latency values of supergraphs that don't use
{\it s-t swaps} are 3, respectively 2.86 times larger, which shows how
poor can be, at times, the degree of available parallelism if the source and
sink are not properly swapped.

\vspace*{-0.15in}
\subsection{Impact of Seed Supergraphs}\label{sec:seed_impact}
\vspace*{-0.15in}

Using seed supergraphs (see \S\ref{sec:supergraphs}) should achieve better
usage of the underlying hardware. We assessed the performance of NVIDIA's push
relabel implementation on seed supergraphs when varying the number of
seeds. Table \ref{tab:seed_impact} shows the results on a Z840 station for two
seeds (four seed supergraphs have shown only marginally better figures). The
comparison to one seed supergraph figures (table \ref{tab:basicGPU}) proves
roughly 10\% improvements, on average.

\vspace*{-0.1in}
\begin{table}[htb]
\centering
\begin{tabular}{ccccc}
\hline
\hline
Time & Min & Avg & Max \\
\hline
\hline
Tesla K40 & 8.82 s & 29.78 s & 54.90 s &   \\
\hline
Titan Black & 7.04 s & 23.49 s & 40.63 s &   \\
\hline
2 x Titan Black & 4.03 s & 12.79 s & 22.45 s &   \\
\hline
\end{tabular}
\vspace*{-0.1in}
\caption{\textrm{Impact of seed supergraphs (2 seeds).}}
\label{tab:seed_impact}
\vspace*{-0.2in}
\end{table}

\vspace*{-0.08in}
\subsection{RPC Performance}\label{sec:rpc}
\vspace*{-0.1in}

We also conducted experiments to evaluate the performance of our RPC version
of the NVIDIA {\it nppiGraphcut} call to access remote GPU boards. The
experiments have used a Tesla K40 board, Z840 workstations and
$\lambda$-schedules of 20 values. The first row of table \ref{tab:rpc} presents
the graph cut times to get the figure-ground segment hypotheses when
using $\lambda$-supergraphs only. We also evaluated the optimizations discussed
in \S\ref{sec:comm} by running larger supergraphs (two seed supergraphs) to
make a better usage of the available network. One can see that the difference
between one and two seed performance in table \ref{tab:rpc} is larger than it
is for local Tesla K40 calls (see tables \ref{tab:basicGPU} and
\ref{tab:seed_impact}). This difference can be accounted to better network
usage in the case of larger supergraphs. The last rows of table \ref{tab:rpc}
show that multi-threaded RPC servers improve the performance when two
consecutive GPU RPCs are submitted to the same server in order to overlap
communication with computation.

\vspace*{-0.1in}
\begin{table}[htb]
\centering
\begin{tabular}{ccccc}
\hline
\hline
Time & Min & Avg & Max \\
\hline
\hline
1 seed & 17.28 s & 48.14 s & 74.99 s &   \\
\hline
2 seeds & 13.91 s & 42.71 s & 72.29 s &   \\
\hline
MT server (1 seed) & 14.59 s & 39.14 s & 62.51 s &   \\
\hline
MT server (2 seeds) & 11.14 s & 33.87 s & 59.82 s &   \\
\hline
\end{tabular}
\caption{\textrm{Performance of remote GPU RPCs.}}
\label{tab:rpc}
\vspace*{-0.1in}
\end{table}

%
%

\vspace*{-0.1in}
\subsection{Parametric Max Flow using Supergraphs and Clusters of GPUs}\label{sec:cluster5}
\vspace*{-0.1in}

Once the performance of the individual boards and mechanisms has been assessed,
we proceeded to set up a cluster of GPUs that would act as a parallel,
cluster-wide parametric maximum flow solver based on supergraphs and the
insights gained from the previous experiments. Thus, we decided to use the
Z840 workstation equipped with two Titan Black cards as a master node, since
these cards are faster than the K40s and local access to them should be faster
than by means of RPC. The other three stations, two Z840 equipped with one
Tesla K40 card each and the Z420 machine hosting the third Titan Black board
have been used to build the cluster of GPU servers. We tested with
$\lambda$-schedules with 20 values and two-seed supergraphs, as using such
supergraphs has shown the best performance. We varied the number of boards
in the cluster from three to five and compared the results with those of CPMC
using the pseudoflow solver on a 6-core Z840 machine. The figures are shown in
table \ref{tab:cluster5} and reveal better overall times for the graph cuts of
the figure-ground segmentation stage based on clusters of GPUs. In terms of
average times, the GPU cluster solutions take roughly 72\%, 62\% and 55\%,
respectively, of the time needed to run the trivially parallel solution using
the pseudoflow algorithm on a 6-core processor.

\vspace*{-0.1in}
\begin{table}[htb]
\centering
\begin{tabular}{ccccc}
\hline
\hline
Time & Min & Avg & Max \\
\hline
\hline
Pseudoflow 6 threads & 4.10 s & 14.35 s & 21.48 s &   \\
\hline
2 Titan Black + 1 K40 & 3.49 s & 10.34 s & 17.86 s &   \\
\hline
2 Titan Black + 2 K40 & 2.91 s & 8.88 s & 15.04 s &   \\
\hline
3 Titan Black + 2 K40 & 2.67 s & 7.85 s & 12.16 s &   \\
\hline
\end{tabular}
\caption{\textrm{The performance of GPU clusters vs. the multi-core based solution.}}
\label{tab:cluster5}
\vspace*{-0.2in}
\end{table}

\vspace*{-0.2in}
\subsection{Graph Cut Accuracy}
\vspace*{-0.15in}

So far, our evaluations concerned running times, but CPMC and image
segmentation algorithms in general need also to fulfill their primary goal of
accuracy. The accuracy of an image segmentation algorithm is
influenced by several factors, the graph cut calculation accuracy
being an important one. Therefore we proceeded to an evaluation of the
performance of the previously tested algorithms also in terms of accuracy.

An image segmentation accuracy measure is a similarity measure,
defined according to the VOC challenge rules \cite{VOC} as the
degree of overlap between the set of segments (pixel masks) S resulted from
the image segmentation algorithm and the ground truth G (correct image segmentation
masks delineated by hand, provided for reference). Alternative measures
include the {\it F-measure} \cite{Alpert}. The overlap measure is computed as follows:
\begin{equation*}
\textrm{Overlap}(S,G) = {| S \cap G | \over  | S \cup G |}
\end{equation*}

The results are presented in table \ref{tab:overlap}. Note that the overlap measure quantifies the accuracy of the whole method (so far, we have
presented running times for graph cuts in the figure-ground segmentation stage
of CPMC).

The difference between the accuracy of CPMC running the
push-relabel algorithm on supergraphs mapped on a cluster of GPU boards to that
of the pseudoflow algorithm is around 1\% and can be probably
accounted to the fact that a pseudoflow algorithm is slightly less precise than a push-relabel algorithm. The point of this
experiment is to show, however, that the NVIDIA implementation can provide reliable, high
quality segments for CPMC. The comparison to the pseudoflow figure ascertains
that.

\vspace*{-0.1in}
\begin{table}[htb]
\centering
\begin{tabular}{cccc}
\hline
  & \multicolumn{1}{c}{Pseudoflow} & \multicolumn{1}{c}{Push-relabel} \\
  & & on GPU cluster\\
\hline
\hline
Avg. overlap (20 $\lambda$) & 0.734 & 0.743 & \\
\hline
\end{tabular}
\caption{\textrm{Image segmentation performance of CPMC in terms of intersection over union overlap.}}
\label{tab:overlap}
\end{table}


\vspace*{-0.41in}
\subsection{Segmentation Accuracy vs. Speed Trade-Off}\label{sec:cluster5-10}
\vspace*{-0.11in}

The CPMC release \cite{cpmc-release} sets the default $\lambda$-schedule size
to 20 values, as this choice has proven to yield the best results in the VOC
challenges \cite{VOC}. However, one can always trade off segmentation accuracy
for improved running times. In this section, we present the results of halving
the size of the $\lambda$-schedule (see table \ref{tab:cluster5-10}). We have
repeated the experiments in \S\ref{sec:cluster5} using two seed supergraphs and
10 $\lambda$ schedules. The comparison to tables \ref{tab:cluster5} and
\ref{tab:seed_impact} (for 2 x Titan Black) shows
a reduction of roughly 40\% of the average running times (slightly more for
the setup with two local Titan Black boards). Using four seed supergraphs
improves only marginally the running times (e.g., for the 5 GPU cluster the
average execution time amounts to 4.54 seconds). It is also worth noting that
the trivially parallel pseudoflow solution using 10 $\lambda$ schedules and six
parfor threads doesn't yield this kind of improvement over the 20
$\lambda$ schedule case (roughly only 20\% decrease).

\vspace*{-0.1in}
\begin{table}[htb]
\centering
\begin{tabular}{ccccc}
\hline
\hline
Time & Min & Avg & Max \\
\hline
\hline
Pseudoflow 6 threads & 3.79 s & 11.36 s & 16.21 s &   \\
\hline
2 Titan Black & 2.46 s & 7.32 s & 12.84 s &   \\
\hline
2 Titan Black + 1 K40 & 2.20 s & 6.05 s & 12.03 s &   \\
\hline
2 Titan Black + 2 K40 & 1.61 s & 5.26 s & 8.34 s &   \\
\hline
3 Titan Black + 2 K40 & 1.68 s & 4.78 s & 7.63 s &   \\
\hline
\end{tabular}
\caption{\textrm{The performance of GPU clusters vs. the multi-core based solution for a 10 value $\lambda$-schedule.}}
\label{tab:cluster5-10}
\vspace*{-0.2in}
\end{table}

\vspace*{-0.15in}
\begin{table}[htb]
\centering
\begin{tabular}{cccc}
\hline
& \multicolumn{1}{c}{Pseudoflow} & \multicolumn{1}{c}{Push-relabel} \\
& & on GPU cluster\\
\hline
\hline
Avg. overlap (10 $\lambda$s) & 0.719 & 0.732 & \\
\hline
\end{tabular}
\caption{\textrm{Image segmentation accuracy for 10 value $\lambda$-schedules.}}
\label{tab:overlap-10}
\end{table}

\vspace*{-0.25in}
The gain becomes even more important if one considers table \ref{tab:overlap-10} that depicts the segmentation accuracy (all of the GPU-based solutions yield
at least 0.732 overlap, so we reported just one figure). Note that,
even if the accuracy of the segmentation drops by roughly 1\% compared to that
reported in table \ref{tab:overlap}, it practically equals that of the
pseudoflow algorithm using twice as many $\lambda$ values. Thus, one can
use fewer $\lambda$ values and get significantly improved running times at
almost no accuracy loss.

\vspace*{-0.15in}
\subsection{Scheduling}\label{sec:scheduling}
\vspace*{-0.15in}

{\bf Static vs. Dynamic Scheduling} As pointed out in \S\ref{sec:sched}, we schedule parametric max flow
computations on supergraphs using a FIFO ordered, dynamically managed
list of computing servers. In contrast, a static solution would assign each
iteration of a {\it parfor} MATLAB loop computing parametric max flows to a
given server in the list.
We instrumented four experiments using {\it parfor} loops with up to five
threads and all of the GPU board combinations from \S\ref{sec:cluster5}.
Each {\it parfor} iteration was statically scheduled {\it mod n}, where {\it n}
was the number of threads/boards used, to run
supergraph max flows on the GPU boards. We used two seed supergraphs
and $\lambda$-schedules of 20 values. Table \ref{tab:scheduling} shows the
results. The comparison to the dynamic solution (tables \ref{tab:cluster5} and
\ref{tab:seed_impact}, for 2 x Titan Black) reveals that dynamic scheduling
improves average times over static solutions by roughly 10\%, 32\%, 34\% and
30\%, respectively. 

\vspace*{-0.1in}
\begin{table}[htb]
\centering
\begin{tabular}{ccccc}
\hline
\hline
Time & Min & Avg & Max \\
\hline
\hline
2 x Titan B. parfor & 4.58 s & 15.48 s & 27.68 s &   \\
\hline
2 Titan B. + 1 K40 parfor & 4.70 s & 15.25 s & 30.11 s &   \\
\hline
2 Titan B. + 2 K40 parfor & 4.14 s & 13.42 s & 27.98 s &   \\
\hline
2 Titan B. + 3 K40 parfor & 3.57 s & 11.24 s & 18.90 s &   \\
\hline
\end{tabular}
\vspace*{-0.1in}
\caption{\textrm{Static scheduling.}}
\label{tab:scheduling}
\end{table}

\vspace*{-0.1in}
{\bf Scheduling Efficiency} Since finding the optimal schedule is proven to be
NP-hard (see \S\ref{sec:sched}), we aimed to find how much worse performs our
dynamic scheduler compared to a theoretically superior algorithm such as LPT
(see \S\ref{sec:sched}). To match the simplest paradigm of parallel,
non-preemptive scheduling for makespan minimization, namely that of Parallel
and Identical Machines \cite{HBookS}, we gathered the running times of our
one seed supergraph cuts for the Z840 machine equipped with two identical Titan
Black boards reported in the last row of table \ref{tab:basicGPU}. We sorted the values in
non-increasing order and applied the LPT algorithm offline. The results have
shown that LPT makespans are 1.7\% smaller, on average, than those of our
dynamic scheduler (the minimal difference among the 500 images being 0.01\% and
the maximum 8.8\%). These results show the efficiency and utility of
our scheduler, given that LPT used a priori known information and thus cannot
be run online, in our case.

\vspace*{-0.15in}
\section{Conclusions}\label{sec:conclusions}
\vspace*{-0.15in}

In this paper we have presented a solution to approximate parallel parametric
maximum flow behavior for image segmentation problems based on supergraphs. We
have also introduced a general, parallel framework that can run parametric maximum flow problems on
various platforms (multi-core, GPU), either locally or distributed in a cluster,
as instructed by a provably efficient dynamic scheduler. 
The framework is also useful as an evaluation tool of the available parallel
maximum flow implementations \cite{Gridcut, NPPI_GC, NPPI_GC_DOC}. We report
the results of using NVIDIA's GPU implementation of the push relabel
maximum flow algorithm together with CPMC \cite{CPMC}, a state-of-the-art image
segmentation algorithm, as a case study that points out the utility of our
framework. The evaluation has shown that our solution achieves near real-time
performance, practically without any segmentation accuracy loss.



\vspace*{-0.1in}
\section*{Acknowledgements}
\vspace*{-0.1in}


This work was supported in part by CNCS-UEFISCDI under CT-ERC-2012-1, PCE-2011-3-0438, and JRP-RO-FR-2014-16. The authors would like to thank NVIDIA Corporation for their generous hardware donation.

\bibliographystyle{plain}
\vspace*{-0.15in}
\bibliography{b1}
\vspace*{-0.15in}


\end{document}